# An Unmanned Aerial Vehicle as Human-Assistant Robotics System

Tejbanta Singh Chingtham[1], G. Sahoo[2], M.K. Ghose[1]

[1]Department of Computer Science & Engineering, Sikkim Manipal Institute of Technology, Rangpo,Sikkim, India
[2]Department of Information Technology, BIT Mesra , Ranchi, India
(chingtham@gmail.com, gsahoo@bitmesra.ac.in, mkghose@smuhmts.edu)

*Abstract*— **According to the American Heritage Dictionary [1], Robotics is the science or study of the technology associated with the design, fabrication, theory, and application of Robots. The term Hoverbot is also often used to refer to sophisticated mechanical devices that are remotely controlled by human beings even though these devices are not autonomous. This paper describes a remotely controlled hoverbot by installing a transmitter and receiver on both sides that is the control computer (PC) and the hoverbot respectively. Data is transmitted as signal or instruction via a infrastructure network which is converted into a command for the hoverbot that operates at a remote site.**

## I. INTRODUCTION

IN practical usage, a mobile robot is an autonomous or semi-autonomous device which performs various tasks either with direct or partial control by human supervision or completely autonomous. Robots are typically used to do tasks that are too dirty or dangerous and that are not suitable for human beings. Industrial Robots used in production process are the most common form of Robots, but are now being recently replaced by consumer Robots for cleaning floors and mowing lawns [2]. Other applications include toxic waste cleanup, underwater and space exploration, surgery, mining, search and rescue, and mine finding. Robots are also finding their ways into entertainment and home health care. This paper lays emphasis on how to control the movement of a Hoverbot using wireless communication. The Hoverbot is installed at remote location which is equipped with transmitter or receiver subsystem. The control commands passed through the computer port will be transmitted to the Hoverbot model through a transmitter or receiver subsystem which in turn guides the model to perform certain task.  The different modules of the system are as follows:

- The first module consists of programming the parallel port. The first step is to enable the port to standard mode. Then the parallel port is tested by interfacing a simple circuit which consists of a LED with the parallel port. The port is programmed such that the LED in the circuit glows on and off when the commands are given when connected to the port.
- The second phase is to prepare the Hoverbot. After the Hoverbot is prepared it is fitted with the Radio Frequency transmitter. The parallel port is interfaced with the Radio Frequency transmitter. The control passed through the port will be transmitted to the model through the Radio

Frequency transmitter which in turn will guide the Hoverbot to perform specified tasks.
- The third module is to test the interfacing of the device with the port and transferring the control signals through the radio transmitter to the Hoverbot and check its functioning [3, 11].
- The fourth phase is to control the Hoverbot through an IIS Server.

## II. PROBLEM DEFINITION

Surveillance of remote and inaccessible areas may require an automated system which can collect information (image or video) of such location and can be transmitted to a team of experts (e.g, disaster management team) for providing real time solutions and remedial actions with drastically reduced human effort and cost with minimum risk of loss of life.
The problem at hand could be divided into three basic parts:
The first part concerns with the program to access the Parallel Port independent of OS in the PC connected to Hoverbot. The second part concerns with development of Hoverbot. The third part deals with the Control of the Hoverbot through internet.
We can use any of the widely available technology such as internet or GPS. But as we know internet is cheaper compared to GPS etc in respect of hardware and software. The basic aim of this project is to make the utility of such automated system available at low cost and sharing of knowledge gained from such experiences stored on a server and can be used on the world wide web [10].

## III. DEVELOPMENT, OPERATING AND MAINTENANCE ENVIRONMENT

The following the operational specifications in term of hardware, Software and Functional requirements [4].
**Hardware:**
- Intel Pentium 4 processor
- RAM: 128 MB (minimum) 256 MB (recommended)
- At least one free parallel port through which the movement signals are transmitted.
- Minimum of 50 MB of free Hard disk space
- Radio freq. transmitter (47MHz)
- Hoverbot.

**Functional Requirements:**
- Get access to the pc–station to which the Hoverbot is connected.
- Send signal to perform the movement of the Hoverbot to the remote pc.
- Signal converted to the radio frequency and transmitted to the Hoverbot.

The functional requirements may be elaborated as follows:

R1.1: Select the Remote pc –station:
• Input: Pc-Station address.
• Output: User prompt to enter the movements.

R1.2: Get Required Movement:
• Input: Only Start, Ready, Fly, Left and Right.
• Output: The requested operation is send to the remote PC-station connected either through LAN or Internet.

R1.3: Transmit the Signal to Hoverbot.
• Input: Radio Frequency signal to the Hoverbot.
• Output: Functioning of the device according to the signals given.

**Non Functional Requirement:**

• Range: Movement of the Hoverbot is about 15 meter high from the PC-station.
• Human Computer Interface: Human Computer Interface is based on the learning provided by MS SAPI.

**Foreseeable Modifications and Enhancements:**
• The movement range or the mobile device may be increased by increasing the radio frequency of the transmitter.
• Better control of the movement of the device.

IV. DESIGN SPECIFICATIONS

Fig 1 describes the various transition state [4] of the Hoverbot. After it is started and can take at least six different states viz., Left, Right, Ready, Start, Fly and Stop.

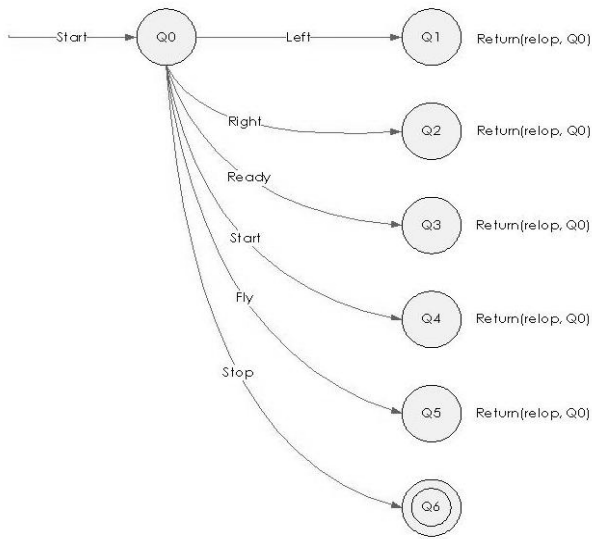

Fig 1 Transition diagram of the various states of the Hoverbot

In Fig 2, the flow sequence of action that the Hoverbot undergoes as it receives each command is described in the form of a flow chart.

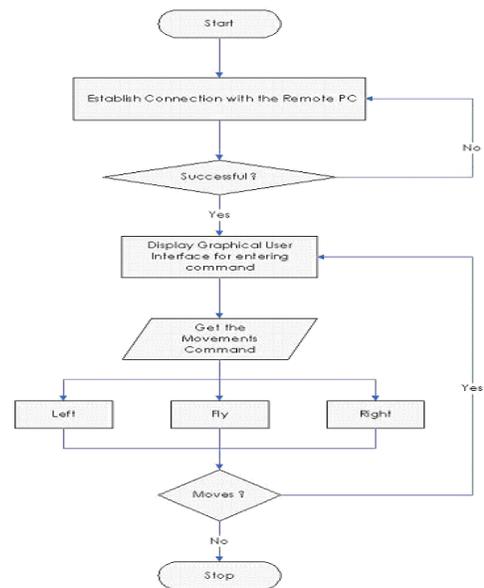

Fig 2 Flowchart depicting action sequence of the Hoverbot

V. HARDWARE IMPLEMENTATION DETAILS

In Fig 3, the Radio Transmitter block diagram is elaborated. The transmitter is supplied with an AC source which in turn drives the input section and also amplified. The Radio Signal is generated and produced as output Radio Frequency which will be transmitted from the tower connected to the PC to the Hoverbot [3]. Fig 4, shows a detailed circuit diagram of the Radio Frequency transmitter to control the Hoverbot is depicted.

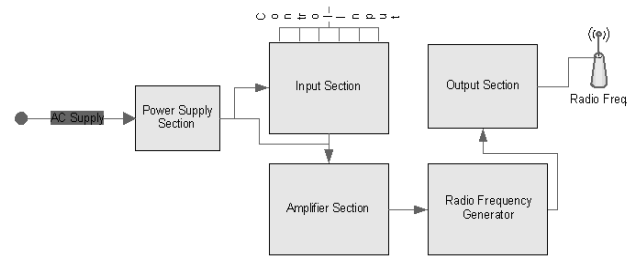

Fig 3 Different Sections of Radio Transmitter [3]

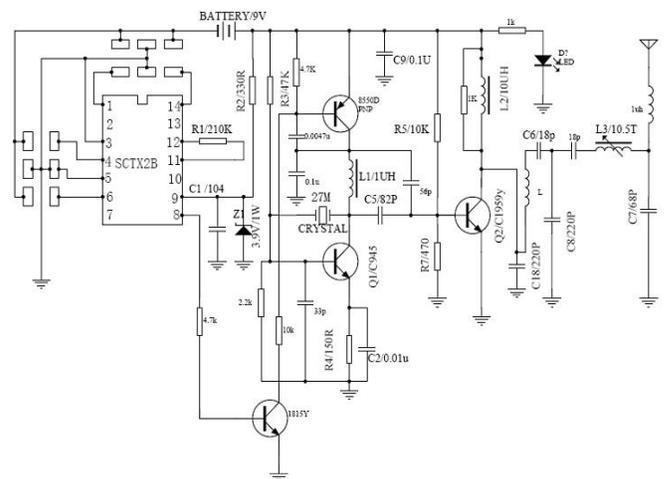

Fig 4 Circuit Diagram for the Radio Frequency Transmission [3,5,6]

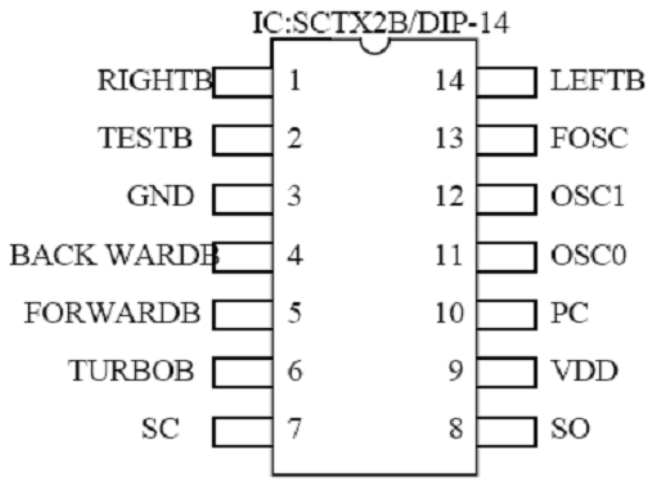

Fig 5 PIN diagram of IC TX-2B [5]

The pin configuration diagram [5] which acts as the heart of the Hoverbot is described in Fig 5.

## VI. CHARACTERISTICS OF IC TX-2B

The characteristics of the main IC used on the Hoverbot IC TX-2B are as follows:

• Working voltage scope: 2.4V~5.0V
• The quiescent current is low.
• May apply to mini remote control and so on compact car, motorcycle, slippery animal-drawn cart, etc.
• SCTX2B/RX2FS [5] is a pair of CMOS integration chip, specially designs uses in to control remotely the vehicle application aspect SCTX2B/RX2FS to have 5 control keys to use in to control controls remotely the vehicle the movement (for example advance, backward right extension, counterclockwise and revolves)

| Parameter | Mark | Minimum value | Typical | Maximum value |
|---|---|---|---|---|
| Working voltage | VDD | 2.4V | 4V | 5V |
| operating current | Idd | 0.5mA | - | 1mA |
| quiescent current | Istb | - | - | 3μA |
| DC O/P actuation electric current | Idrive | 2.5mA | - | - |
| AC O/P actuation electric current | Idrive | 2.5mA | - | - |
| AC O/P frequency | Faudio | 500Hz | - | 1KHz |

Table 1 Parameter Characteristics of IC TX-2B [5]

## VII. CONFIGURING THE PARALLEL PORT

Parallel ports are most often used by microprocessors to communicate with peripherals. The most common kind of parallel port is a printer port, e.g. a Centronics port which transfers eight bits at a time. Disks are also connected via special parallel ports, e.g. SCSI, ATA. The parallel port of an IBM-PC compatible is one of the standard computer port that brings standard computer logic voltages directly out to a set of pins. It is much beloved by experimenters and engineers who often use it for inexpensive computer controlled projects. Standard logic voltages are virtually harmless: five volts (roughly the same as two run-down flashlight batteries), and ground (zero volts) [6].

Parallel ports have four types of pins [6]:

**Data pins:** usually 8, sometimes 16, and sometimes with an extra pin for a parity bit. It can either be unidirectional (e.g., from a computer to a printer) or bidirectional.

**Control pins:** It is used to send control signals such as STROBE to indicate that the data on the data pins is ready and R/W to specify whether bidirectional ports are reading or writing data.

**Status pins:** It is used to send status signals such as BUSY to indicate the device is not ready to receive data and ACK to acknowledge successful receipt of the symbol.

**Ground pins:** It is used to complete the circuits from the other pins.

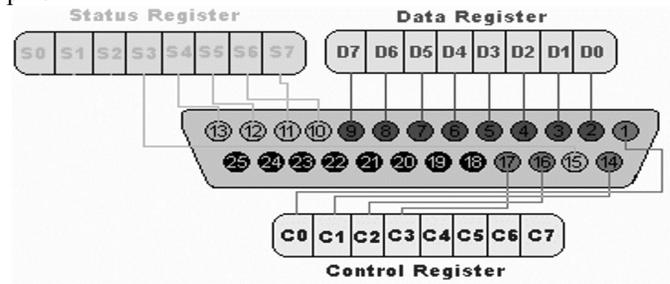

Fig 6 A Pin configuration of a typical Parallel Port [6]

## VIII. RADIO FREQUENCY SPECTRUM AND ITS UTILITY

Radio frequency, or RF, refers to that portion of the electromagnetic spectrum in which electromagnetic waves can be generated by alternating current fed to an antenna [7, 11]. Such frequencies account for the following parts of the spectrum shown in the table below:

| Band name | Abb | ITU band | Frequency Wavelength | Example uses |
|---|---|---|---|---|
| Extremely low frequency | ELF | 1 | 3–30 Hz 100,000 km – 10,000 km | - |
| Super low frequency | SLF | 2 | 30–300 Hz 10,000 km – 1000 km | Communication with submarines |
| Ultra low frequency | ULF | 3 | 300–3000 Hz 1000 km – 100 km | - |
| Very low frequency | VLF | 4 | 3–30 kHz 100 km – 10 km | Submarine communication, avalanche beacons, wireless |

| | | | | heart rate monitors |
|---|---|---|---|---|
| Low frequency | LF | 5 | 30–300 kHz 10 km – 1 km | Navigation, time signals, AM long wave broadcasting |
| Medium frequency | MF | 6 | 300–3000 kHz 1 km – 100 m | AM broadcasts |
| High frequency | HF | 7 | 3–30 MHz 100 m – 10 m | Shortwave broadcasts and amateur radio |
| Very high frequency | VHF | 8 | 30–300 MHz 10 m – 1 m | FM and television broadcasts |
| Ultra high frequency | UHF | 9 | 300–3000 MHz 1 m – 100 mm | television broadcasts, mobile phones, wireless LAN |
| Super high frequency | SHF | 10 | 3–30 GHz 100 mm – 10 mm | microwave devices, mobile phones (W-CDMA), WLAN, most modern Radars |
| Extremely high frequency | EHF | 11 | 30–300 GHz 10 mm – 1 mm | Radio astronomy, high-speed microwave radio relay |

Table 2 RF Spectrum table [7]

In our case we are using a 49 MHz crystal to generate radio signals through a remote control so that it can easily take off the Hoverbot at the height of 30 meters.

## IX. SOFTWARE IMPLEMENTATION DETAILS

Pseudo-code for Functions and Event-Handlers [8] for the implementation to run the Hoverbot are described as under. The Implementation was coded using Microsoft Visual Studio 6.

**Event:** *Form_load ()*
**Called:** When the system starts.
**Global parameters used:** URL
**Local parameters used:** None

Begin
Initialize Server.
If (server down)
Begin
    Display error message.

End
Else
Begin
    Display the control form.
End
End
**Function:** *Portaccess (databits)*
**Called:** By events Left_button_click (), Right_button_click (), Start_button_click (), Ready_button_click (), Fly_button_click ()
**Calls**: Inpout32.dll
**Global parameters used:** Port Address
**Local parameters used:** None

Begin
    Transmit data bits to the Port.
End

**Event:** *Left_button_click ()*
**Called:** When the left button is clicked.
**Calls:** PortAccess ().
**Global parameters used:** None.
**Local parameters used:** Data_bits
Begin
    Call PortAccess (Data_bits)
End

**Event:** *Right_button_click ()*
**Called:** When the right button is clicked
**Calls:** PortAccess ()
**Global parameters used:** None.
**Local parameters used:** Data_bits
Begin
    Call PortAccess (Data_bits)
End

**Event:** *Stop_button_click ()*
**Called:** When the stop button is clicked
**Calls:** PortAccess ()
**Global parameters used:** None.
**Local parameters used:** Data_bits
Begin
    Call PortAccess (Data_bits)
End

**Event:** *Fly_button_click ()*
**Called:** When the fly button is clicked
**Calls:** PortAccess ()
**Global parameters used:** None.
**Local parameters used:** Data_bits
Begin
    Call PortAccess (Data_bits)
End

**Event:** *Start_button_click ()*
**Called:** When the start button is clicked
**Calls:** PortAccess ()
**Global parameters used:** None.
**Local parameters used:** Data_bits
Begin

    Call PortAccess (Data_bits)
End

**Event:** *Ready_button_click ()*
**Called:** When the ready button is clicked
**Calls:** PortAccess ()
**Global parameters used:** None.
**Local parameters used:** Data_bits
Begin
    Call PortAccess (Data_bits)
End

## X. RESULTS

Implementation of the above code produces the following results. Fig 7 through 15 is the snapshot of different controls of the Hoverbot and its operation on receiving the commands [9].

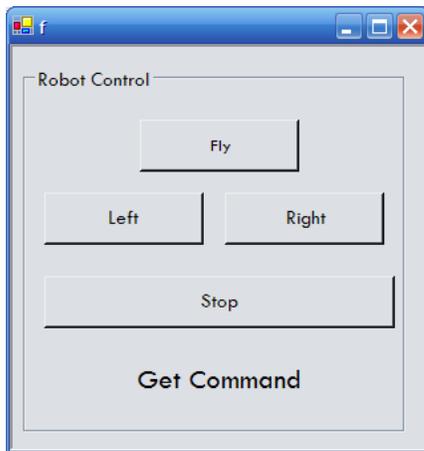
Fig 7 Waiting for Command

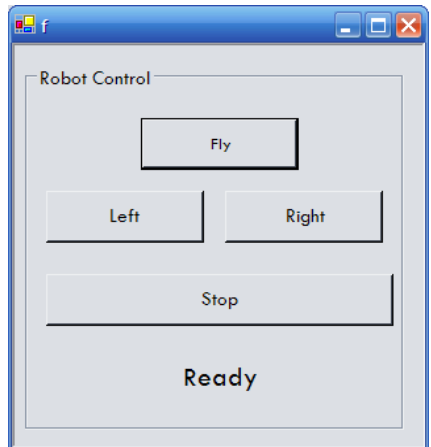
Fig 8 After Clicking on Fly the Hoverbot is ready to take off

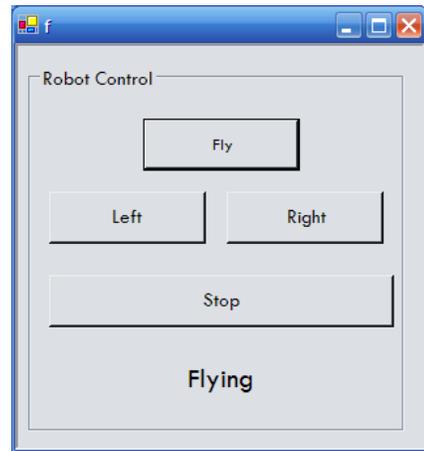
Fig 9 Clicking on Fly again instructs Hoverbot to fly

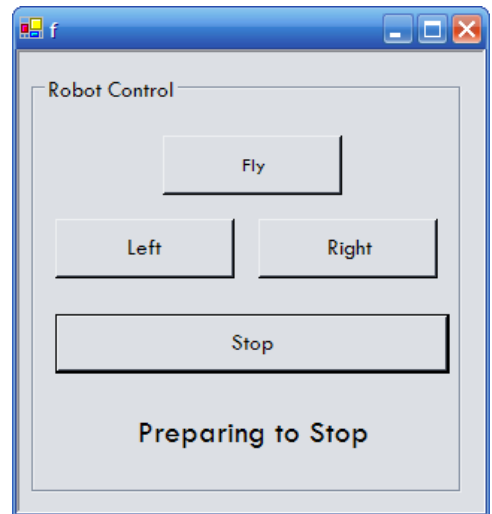
Fig 10 Stop Instruction

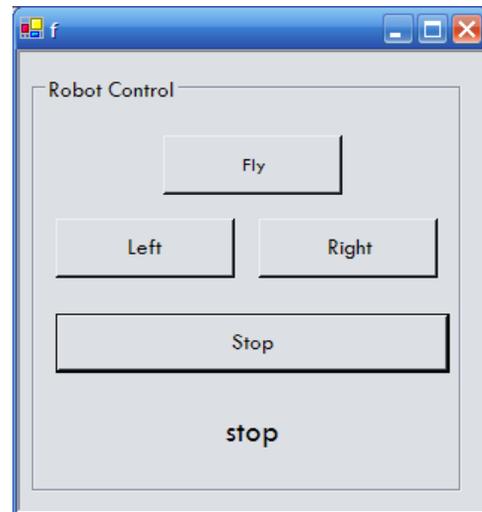
Fig 11 Hoverbot application halts the Hoverbot

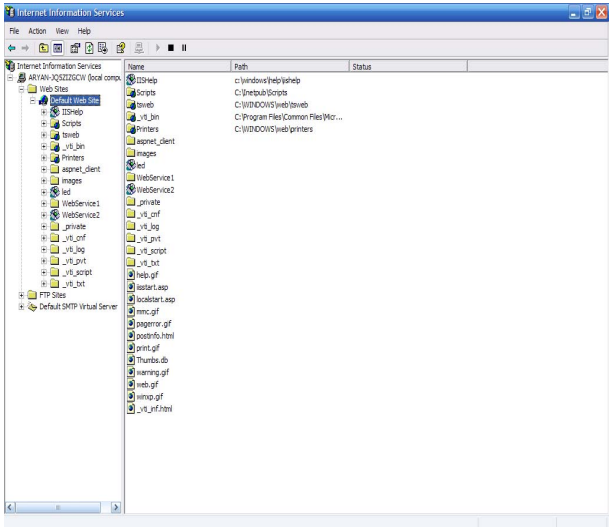

Fig 12 Server Side Control Files Interfaces

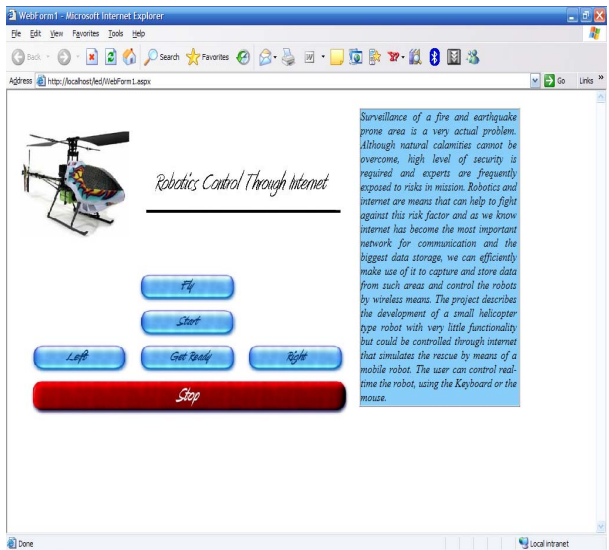

Fig 13 Client Side Interface [11]

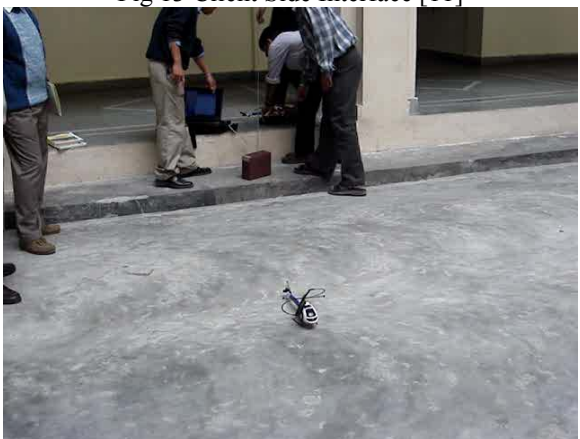

Fig 14 Hoverbot at station [11]

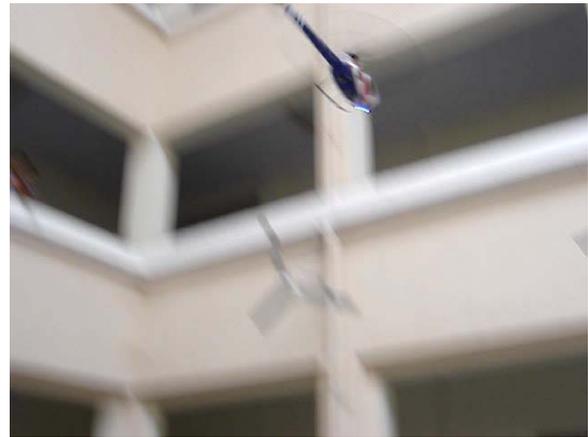

Fig 15 Hoverbot in air [11]

ACKNOWLEDGMENT

This work is supported by AICTE-RPS grant vide Grant No 8023/BOR/RID/RPS-235/2008-09.